\documentclass{article}

\usepackage{latexsym}
\usepackage[T1]{fontenc}
\usepackage[utf8]{inputenc}
\usepackage{microtype}
\usepackage{inconsolata}
\usepackage{graphicx}
\usepackage{appendix}
\usepackage{setspace}
\usepackage{hyperref}
\usepackage{multirow}

\usepackage[final]{neurips_2024}

\usepackage[utf8]{inputenc} 
\usepackage[T1]{fontenc}    
\usepackage{hyperref}       
\usepackage{url}            
\usepackage{booktabs}       
\usepackage{amsfonts}       
\usepackage{nicefrac}       
\usepackage{microtype}      
\usepackage{xcolor}         

\title{Simulating User Agents for Embodied Conversational-AI}

\author{
    Daniel Philipov, Vardhan Dongre, Gokhan Tur, Dilek Hakkani-Tür \\
    University of Illinois at Urbana-Champaign \\
    \texttt{\{dp33, vdongre2, gokhan, dilekh\}@illinois.edu}
}

\begin{document}

\maketitle
\begin{abstract}
Embodied agents designed to assist users with tasks must possess the ability to engage in natural language interactions, interpret user instructions, execute actions to complete tasks and communicate effectively to resolve issues. However, collecting large-scale, diverse datasets of situated human-robot dialogues to train and evaluate such agents is expensive, labor-intensive, and time-consuming. To address this challenge, we propose building a large language model (LLM)-based user agent that can simulate user behavior during interactions with an embodied agent in a virtual environment. Given a specific user goal (e.g., make breakfast), at each time step during an interaction with an embodied agent (or a robot), the user agent may "observe" the robot actions or "speak" to either proactively intervene with the robot behavior or reactively answer the robot’s questions. Such a user agent assists in improving the scalability and efficiency of embodied dialogues dataset generation and is critical for enhancing and evaluating the robot's interaction and task completion ability, as well as for future research, such as reinforcement learning using AI feedback. We evaluate our user agent's ability to generate human-like behaviors by comparing its simulated dialogues with the benchmark TEACh dataset. We perform three experiments: zero-shot prompting to predict the dialogue act from history, few-shot prompting, and fine-tuning on the TEACh training subset. Our results demonstrate that the LLM-based user agent can achieve an F-measure of 42\% in mimicking human speaking behavior with simple zero-shot prompting and 43.4\% with few-shot prompting. Through fine-tuning, we achieved similar success in deciding when to speak but much greater success in deciding what to say, from 51.1\% to 62.5\%. These findings showcase the feasibility and promise of the proposed approach for assessing and enhancing the effectiveness and reliability of robot task completion through natural language communication.
\end{abstract}

\section{Introduction}

Embodied agents or robots designed to assist users with tasks should be able to engage in natural language interactions and communicate effectively with their users to resolve issues that arise during task completion.
It is costly and labor intensive to collect datasets to train such agents, as also supported by the few datasets available~\cite[among others]{CVDN, alfred, hurdl, Padmakumar2022}. Furthermore, interactive datasets are needed to evaluate the task completion abilities of these agents. 

\begin{figure}[t]
\centering
\includegraphics[width=0.8\textwidth]{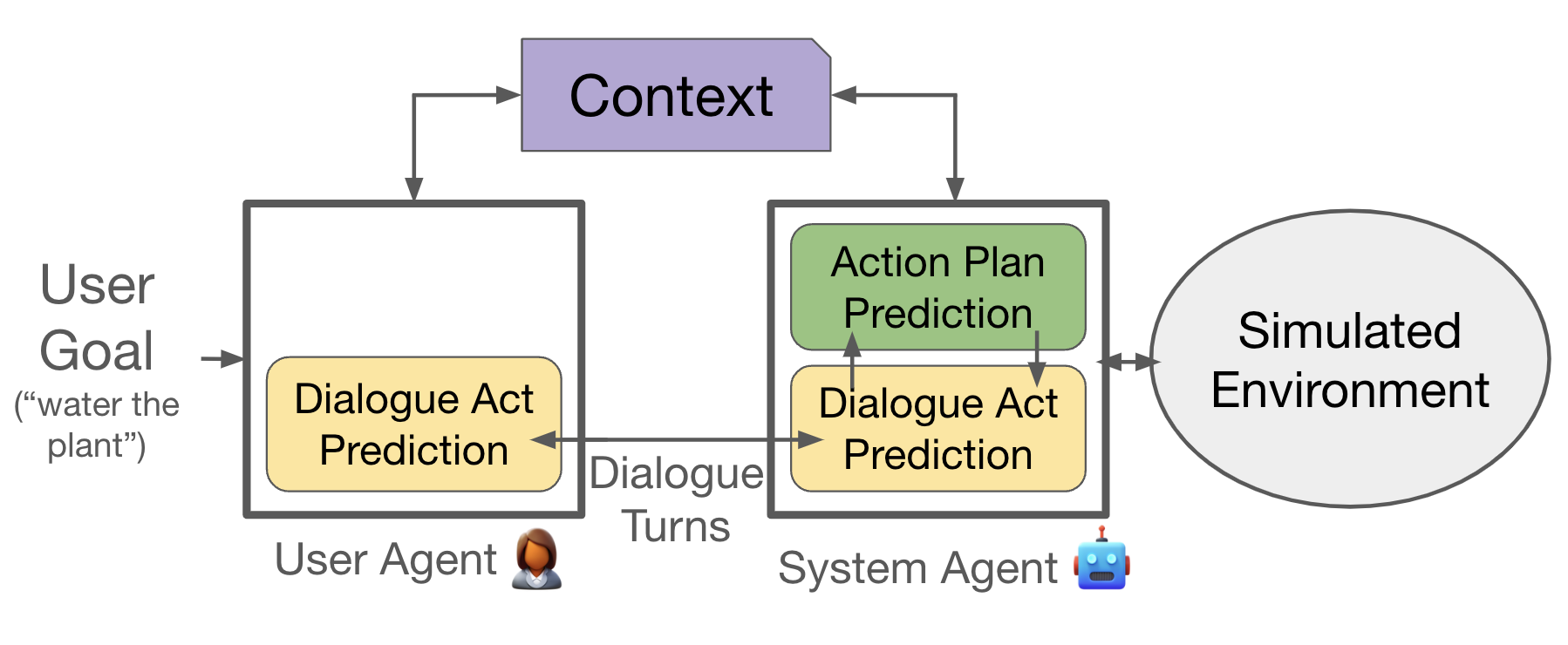}
\vspace*{-3ex}
\caption{A depiction of the framework that includes a user simulator interacting with an embodied agent to complete a task given as a user goal.}
\label{usersim}
\vspace*{-3ex}
\end{figure}

\begin{figure*}[ht]
\centering
\includegraphics[width=\textwidth]{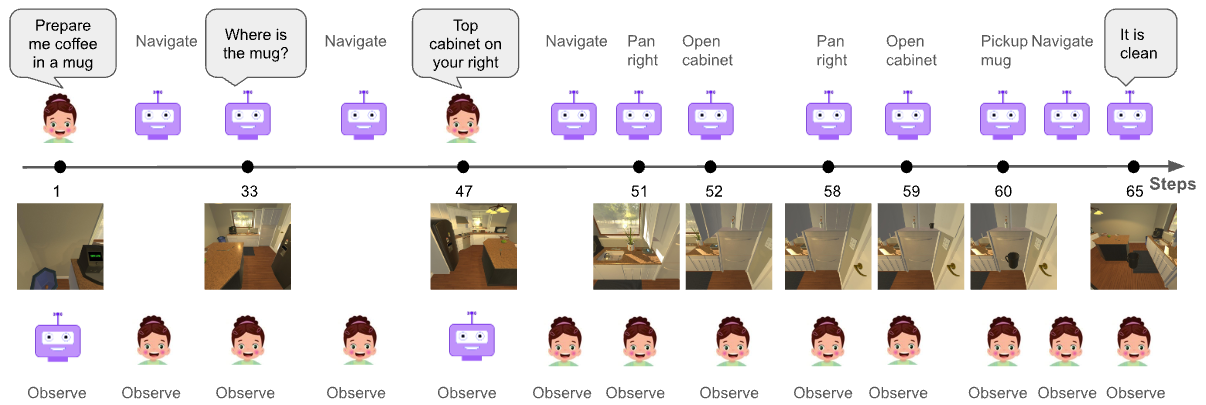}
\vspace*{-2ex}
\caption{A sample session from TEACh dataset. At each step, either the user or the embodied agent takes an action. Images at the bottom show the egocentric views captured by the robot after the action of that time step is executed.}
\label{fig:teach}
\vspace*{-2ex}
\end{figure*}

In this work, we propose an LLM-based user proxy agent that simulates user behavior in human-robot interactions using a virtual environment, AI2Thor~\cite{ai2thor}. 
While the use of user simulators for task-oriented dialogue systems (TODS) is well-established (see Section~\ref{sec:rw}), the application of conversational embodied AI user simulators that leverage LLMs is relatively unexplored. Addressing this gap is significant given the increasing capabilities of LLMs in generating natural and contextually appropriate dialogue. Embodied agents must interact with their users due to various reasons, such as resolving ambiguous requests, seeking further clarification, and requesting confirmation, and hence the user simulator should be able to handle each of these cases, in addition to proactively providing feedback or clarifications to the robot.
Figure~\ref{fig:teach} includes an example of a robot requesting additional information in the TEACh dataset~\cite{Padmakumar2022}.

The simulated user agent can interact with the embodied agent to synthesize dialogue datasets, evaluate the embodied agent and help in refining the embodied agent's abilities for task completion through reinforcement learning using AI. Given a concrete user goal (e.g., {\em make breakfast}), at each time step during the interaction, the user agent may "observe" the robot actions or "speak" in order to either proactively intervene with the robot behavior or reactively respond to the robot when needed.  

In TODS, turn-taking for a user simulator is arguably simpler, as the user agent typically produces a turn after each system turn until the conversation ends. However, in the case of embodied AI, users can produce utterances to instruct the embodied agent, to answer questions the embodied agent asks or to correct, provide feedback for embodied agent's actions, or simply observe the robot actions. 

In our work, we propose prompting an LLM with a user goal and few-shot in-context examples, to predict user actions throughout an interaction session with a robot. 
Figure~\ref{usersim} depicts the overall framework, where the user agent observes the actions of the embodied agent in the environment and its natural language utterances and determines when to speak and what to say in terms of higher level dialogue acts. 

To evaluate the user simulator, we assess its ability to mimic the actual user actions using the TEACh dataset~\cite{Padmakumar2022}. Our results demonstrate the feasibility of our approach to enhance the effectiveness and reliability of human-robot interactions in achieving tasks through natural language communication.

The novel contributions of our work include the development of an innovative framework that predicts user dialog actions in user-robot interactions, which includes determining a user dialog act as well as its appropriate timing and an investigation into the performance of LLMs in executing these tasks. This provides insights into the accuracy of LLMs in simulating realistic user behavior when interacting with embodied agents, thereby enhancing the capabilities of embodied conversational AI systems in real-world scenarios.
\vspace*{-1ex}
\section{Related Work}
\label{sec:rw}
\vspace*{-1ex}
In this work, we represent the actions of the conversational user actions in terms dialogue acts. Annotations of dialogue acts are frequently found in task-oriented dialogue datasets and are often utilized to determine the next action for the agent in dialogue management or the next action for the user in user simulation. \cite{Gella2022} presents a dialogue act annotation schema for embodied task completion utilizing dialogues of the TEACh dataset and then extends it by fine-tuning language models to tag dialogue acts (see Appendix A for a list of dialogue acts used in this work and their explanations), predict the next dialogue act given a dialogue history, and guide the agent’s non-dialogue behavior. Our user simulator approach works similarly, except, in addition to dialogue acts, we also predict the turn taking behavior of the user agent.

Many previous studies proposed methods for building user simulators for TODS~\cite[among others]{schatzmann2009hidden, gur2018user, asri2016sequence}. For instance, \cite{davidson2023user} developed a user simulator using in-context learning with LLMs that generates linguistically diverse and complex utterances similar to natural real-world user inputs to TODS. Using a predefined set of goals aligned with the schema of the target TODS, they used a user simulator for interactive evaluations. 
However, to our knowledge, fewer studies focused on user simulation for embodied agent interactions. \cite{Padmakumar2023} developed a framework for generating synthetic embodied dialogues to train agents. Their dialogues are produced using an agenda-based user simulator~\cite{schatzmann2009hidden, gur2018user, shi-etal-2019-build}, which simulates human conversations through a predefined set of expected user actions. These actions include natural language instructions, task-related information exchange, and environment-based actions. In contrast, our work relies on instructing LLMs to generate synthetic dialogues for this task, offering a more flexible approach.
\cite{gao2022dialfred} employed an oracle-based system to simulate human-like interactions for embodied instruction-following tasks. The oracle provides predefined responses to queries posed by agents. While this user simulator represents a viable initial approach, it may not fully capture the intricacy and heterogeneity inherent in real-world human interactions. In contrast, an LLM-based agent can offer richer, more adaptable simulations.
\cite{dai2023think} simulate users to evaluate embodied agents in interactions. Their framework utilizes a pre-trained LLM to generate user responses based on the given context, including dialogue history and task-related information, but the LLM-based user simulator does not model turn-taking for the user.

\vspace*{-1ex}
\section{Approach}
\vspace*{-1ex}
A session, $S = \{(s_1, a_1), ... , (s_T, a_T)\}$, of interactions between a user and an embodied agent can be represented as a sequence of pairs, $(s_i, a_i)$, where $s_i$ represents the agent that performed an action at step $i$, $s_i \in \{\rm robot, user\}$, and $a_i$ represents the action that the agent $s_i$ performed at that step. 
The set of possible actions $A$ is the union of the set of possible dialogue acts, $D$, for conversational actions that the robot or the user can execute and the set of navigation and object manipulation actions, $P$, that the robot can execute.
Then, building a user simulator involves predicting whether the user should not perform a conversational action and simply "observe" (i.e., the robot is performing an action at the next time step) or whether the user should perform a conversational action and what should be the dialogue act of that action, for each time step, given all the interaction history until that time. Hence, the input to the user simulator for step i, $x_i$, is the sequence $(s_1, a_1), ... , (s_{i-1}, a_{i-1})$, and the goal of simulation is to predict $y_i \in \{{\rm "observe"}\} \cup D$. In this work, similar to ~\cite{Padmakumar2023}, the user simulator produces dialogue acts (e.g., Instruction or Confirm) which can then be converted to natural language user responses using templates formed from the examples of the training dataset.

These steps in a session are not supposed to be regularly distributed over time, and their duration would differ depending on the action. To tackle this issue, if an "observe" action is predicted for the user, we expect the user simulator to wait until the robot performs an action, and after that, user simulator outputs its next action. However, this may result in an infinite loop, in case the robot also doesn't take an action. Hence, during inference, we introduce a maximum time to observe, and then the user simulator is forced to predict a conversational action. In this work, we experiment with zero- and few-shot methods that instruct an LLM to predict the next user action. 

\vspace*{-1ex}
\subsection{Instructing LLMs for Simulating Users}

The first thing presented to the LLM is a description of the role it should play. In the description of the problem, dialogue acts are introduced and explained in order to be guide the LLM in the examples and the answers to the tasks given. Dialogue acts are explained using the descriptions in~\cite{Gella2022} (See \hyperref[app:ex]{Appendix}). 

In experiments labeled "FS," for few-shot, we included five example scenarios. Examples were selected as variable-length sequences from the randomly selected sessions of the training dataset. The examples all began at the beginning of that session, and continue for a random length of turns. Examples were re-drawn from the dataset if too many of the user turns (i.e., greater than 35\%) had the "OBSERVE" answer, in an effort to represent more of the conversational turns to the LLM. In this case, up to two examples can be answered "OBSERVE."

Each example began with the user goal:
\vspace*{-1ex}
\begin{quote}
    {\ttfamily \footnotesize Goal: make me a sandwich}
\end{quote}
\vspace*{-1ex}
Then, continued with a sequence of "COMMANDER" and "DRIVER" action pairs, i.e.: 
\vspace*{-1ex}
\begin{quote}
{\ttfamily \footnotesize
COMMANDER: <observe>\\
DRIVER: <pickup Bread>}
\end{quote}
\vspace*{-1ex}
Non-dialogue actions were surrounded by angle brackets, signifying that they are not text. Actions with a target, such as pickup actions, had it enclosed within the angle brackets. Dialogue actions were written in plain text, followed by the dialogue act(s) enclosed in pairs of angle brackets, i.e.: 
\vspace*{-1ex}
\begin{quote}
{\ttfamily \footnotesize
COMMANDER: i would like the remote put on the side table <<Instruction>>\\
DRIVER: <observe>}
\end{quote}
\vspace*{-1ex}
Each example ended with a request for the user agent's response, followed by the answer to the example problem, formatted like the following:
\vspace*{-1ex}
\begin{quote}
{\ttfamily \footnotesize
COMMANDER response: \\
Instruction}
\end{quote}
\vspace*{-1ex}

Experiments labeled "ZS," for zero-shot, skipped directly from the dialogue acts to the task, and did not include examples.
The task was then described to the LLM, where the LLM was instructed to respond with either "OBSERVE" or a dialogue act. After, the scenario was described to the LLM, in the same format as the examples, except the answer is omitted for the LLM to predict.




\section{Experiments}

\subsection{Dataset}
For experimental validation, we use the TEACh dataset~\cite{Padmakumar2022} of situated dialogues between human annotators playing the role of a user (Commander) or a robot (Follower/Driver).  The annotators interact to complete a high level task from a given set of tasks in simulated household environments, such as making coffee or watering a plant. Each interaction session includes the definition of the high-level task (presented only to the Commander) and the sequence of actions performed by the two sides during that session. The robot engages in a dialogue with the user to learn the task to be completed, obtain information about objects involved in the task, and perform navigation or object manipulation actions in the simulated environment to achieve the task.
The user can observe the actions that the robot is taking in the environment and interact with the robot in natural language. In this work, we use the dialogue acts for user and robot utterances annotated in~\cite{Gella2022}. 

TEACh dataset consists of about 3K interaction sessions split into 5 subsets: training, validation seen, validation unseen, test seen, and test unseen. Since the test sets are not publicly available, we report our results on the validation seen subset. This subset includes 181 sessions and 7923 steps. In these steps, either the user issues a conversational turn (17.6\%), the robot issues a conversational turn (13.2\%), or the robot issues a navigation or object manipulation action (69.2\%).

\subsection{Experiment Details}
Our experiments evaluated the user simulation model’s capability across different prompting methods and configurations to replicate human-like interactions with embodied agents. We tested two primary prompting approaches: zero-shot, where the model predicts actions without prior example guidance, and few-shot, where selected in-context examples help orient the model’s responses. Additionally, to assess the influence of non-verbal actions, we compared model performance on datasets both with and without move actions, examining how these configurations impacted turn-taking (Speak-F1) and Dialogue Act accuracy. A baseline model was established for comparison, reflecting simple reactive turn-taking and majority class assignments for dialogue acts. Fine-tuning was applied to Llama 3.1 8B and RoBERTa-base models over multiple epochs, providing a comparison to zero- and few-shot approaches and offering insight into performance gains achievable through model adaptation.

\subsection{Metrics}
To compute how well our models mimic the actual user behavior, we compute two metrics: the F-measure for the prediction of "speak" turns (Speak-F1) and the F-measure for the prediction of the dialogue actions (DA-F1). Speak-F1 aims to answer the question of whether the simulator can mimic the actual user's behavior to determine when to talk, and DA-F1 aims to answer the question of whether the simulator can mimic the user's behavior to determine what to say, assuming the model has spoken when expected to.

\subsection {Results and Discussion}

\begin{table}[htbp]
    \centering
    \begin{tabular}{l|c|c|c}
    \toprule
    Condition      & Baseline & GPT4-ZS & GPT4-FS \\
    \midrule
    P: R action    & 0.0\%    & 21.3\%  & 23.7\%  \\
    P: R observe   & 0.0\%    & 27.7\%  & 33.4\%  \\
    P: R speak     & 68.4\%   & 78.0\%  & 78.2\%  \\
    \midrule
    Overall        & 37.7\%   & 42.0\%  & \textbf{43.4\%} \\
    \bottomrule
    \end{tabular}
    \vspace{2ex}
    \caption{Speak-F1 results showing the accuracy of predicting when the user should take a turn, broken down by the previous robot (R) action: movement, observation, or speech. The baseline assumes the user simulator only speaks when spoken to, while GPT4-ZS and GPT4-FS represent zero-shot and few-shot GPT-4 models.}
    \label{tab:speak-f1}
\end{table}

In experiments, we first computed the quality of the prediction of conversational user turns with the zero- and few-shot prompting of GPT4, as shown in Table~\ref{tab:speak-f1}. We compared these with a baseline inspired from TODS user simulators, where the user is predicted to speak every time after the robot issues a conversational turn.
In addition to overall performance results, we also breakdown scores according to the previous robot action (e.g., "P: R speak" denotes the Speak-F1 results only after the robot utterances.)
Our results indicate that both methods surpass this baseline and the GPT4 with few-shot examples perform the best, obtaining an Speak-F1 score of 43.4\%.

\begin{table}[h]
    \centering
    \begin{tabular}{l|l|c}
    \toprule
    Approach      & Model         & DA-F1 \\ 
    \midrule
    Instruction-Only & Baseline & 26.53\% \\
    \midrule
    Zero-shot     & GPT4      & 35.15\% \\
                  & Llama 3.1  & 10.7\%  \\ 
    \midrule
    Few-shot      & GPT4       & 51.13\% \\
                  & Llama 3.1  & 12.4\%  \\ 
    \midrule
    Fine-tuned    & Llama 3.1  & 36.70\% \\
                  & RoBERTa-B     & 62.48\% \\
    \bottomrule
    \end{tabular}
    \vspace*{2ex}
    \caption{DA-F1 results for zero-shot, few-shot, and fine-tuned approaches. Zero-shot (ZS) and few-shot (FS) results are shown for GPT-4 and Llama 3.1 8B, while fine-tuned results include Llama 3.1 8B (2 epochs) and RoBERTa-B (10 epochs). The baseline assigns the majority class (Instruction) to all turns. DA-F1 results show the prediction accuracy of user-turn dialogue acts.}
    \label{tab:da-f1-zero-few-fine-tuned}
\end{table}

Table~\ref{tab:da-f1-zero-few-fine-tuned} shows the DA-F1 scores, where a simple baseline of assigning every predicted user turn the majority dialogue act class (i.e., Instruction) has been used. These results seem inferior to the full fine-tuning approach of~\cite{Gella2022} that obtains an F-measure of $59.26\%$ on this subset, however, RoBERTa has the advantage of being limited to these labels. It seemingly outclasses the other models in the DA-F1 measure. This advantage, however, limits the ability to generate a user-response.

\subsection{Dialogue Act level Analysis}
\begin{figure*}[ht]
\centering
\includegraphics[width=\textwidth]{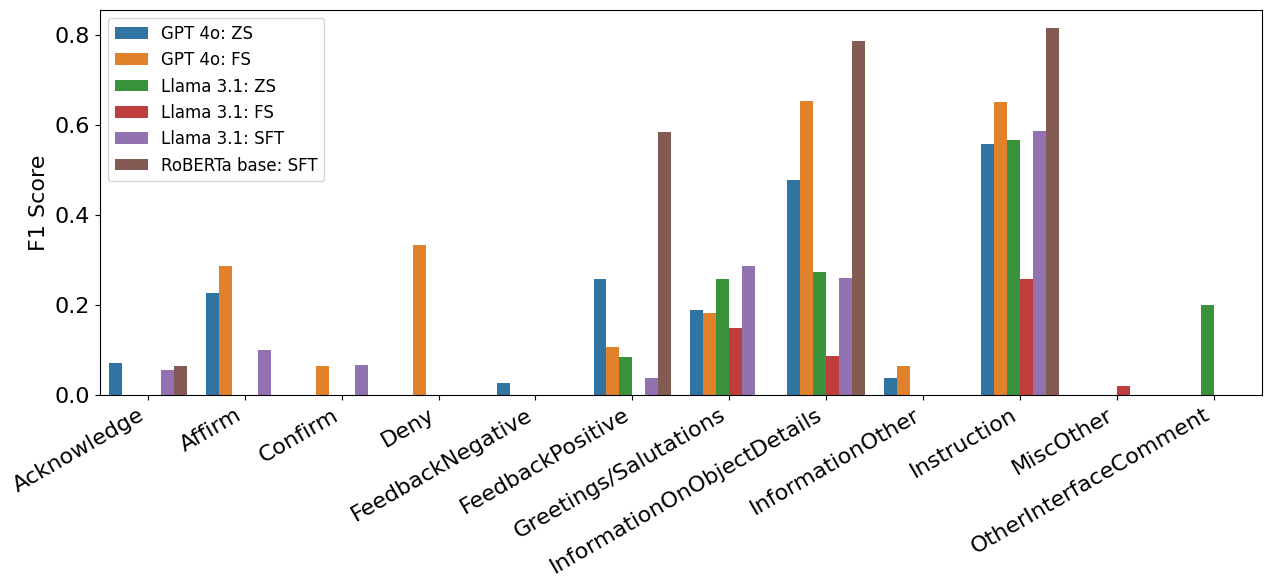}
\caption{Distribution of F-1 Scores across different Dialogue Acts. Robot-only dialogue acts are omitted.}
\label{fig:da-analysis}
\vspace*{-2ex}
\end{figure*}
The distribution of F1 scores across dialogue acts as shown in Figure \ref{fig:da-analysis} reveals notable patterns in the performance of GPT-4 and Llama 3.1 models under zero-shot (ZS) and few-shot (FS) approaches. GPT-4 consistently outperforms Llama 3.1, with its few-shot implementation showing the highest F1 scores across most dialogue acts. This suggests that GPT-4 benefits significantly from task-specific examples. Both models demonstrate strengths in common dialogue acts such as 'Instruction' and 'InformationOnObjectDetails', indicating their proficiency in task-oriented communication. However, there's a marked performance drop for more nuanced or less frequent acts. 'AlternateQuestions', 'RequestForInstruction', and 'RequestForObjectLocationAndOtherDetails' prove particularly challenging, especially for Llama 3.1, highlighting the difficulty in capturing complex query structures. Interestingly, 'Acknowledge' and 'Greetings/Salutations' show high variance across models and training approaches, suggesting that seemingly simple acts can be context-dependent and thus harder to predict consistently. The acts 'RequestMore', 'RequestOtherInfo', and 'OtherInterfaceComment' have near-zero F1 scores for all models, as expected because these dialogue acts are typically reserved for the robot, as shown in Section \ref{app:ex:da-exp}. The zero-shot performance of GPT-4 is notably robust, often rivaling its few-shot counterpart, which underscores its strong pre-training and generalization capabilities. This variance in performance across dialogue acts suggests that future improvements in dialogue systems may benefit from act-specific optimization strategies, particularly focusing on the more challenging and nuanced dialogue acts.

\subsection{Impact of Move Actions}
\begin{figure*}[ht]
\centering
\includegraphics[width=\textwidth]{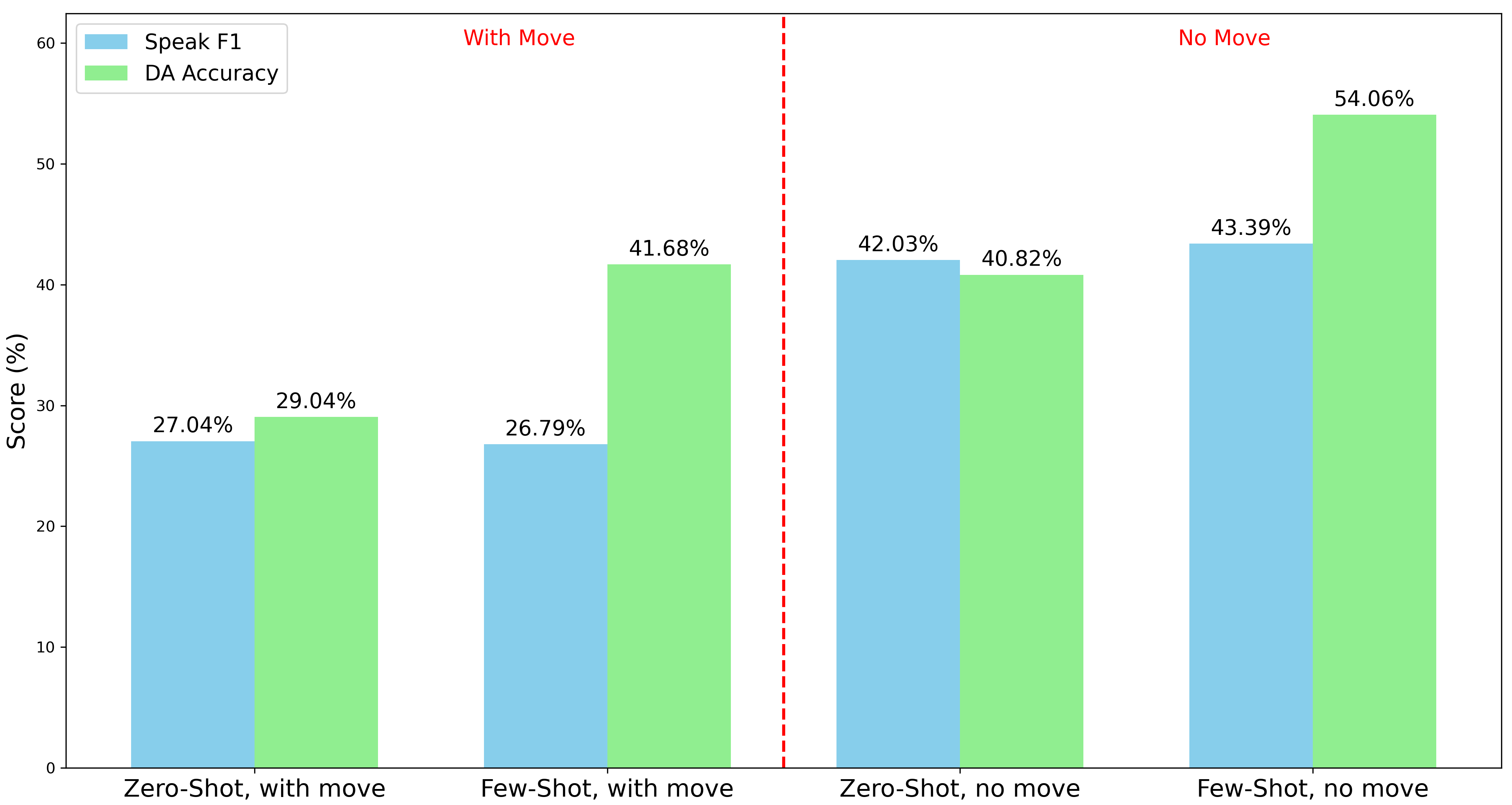}
\caption{Impact of Move Actions on GPT-4 Performance. The figure illustrates the effect of including move actions user behavior modeled by GPT-4 in this case. The results are evaluated using Speak-F1 (blue) and DA Accuracy (green) metrics.  }
\label{fig:move}
\end{figure*}
The impact of move actions on GPT-4's performance is strikingly illustrated in Figure \ref{fig:move}. When move actions are included, the Speak F1 scores remain relatively low and consistent between zero-shot (27.04\%) and few-shot (26.79\%) approaches, indicating that the presence of move actions significantly hampers the model's ability to accurately predict speaking turns. However, the exclusion of move actions leads to a dramatic improvement in Speak F1 scores, jumping to 42.03\% for zero-shot and 43.39\% for few-shot scenarios. This substantial increase suggests that move actions introduce noise that confuses the model's decision-making process for speech prediction. The impact on Dialogue Act (DA) Accuracy is even more pronounced, with few-shot learning showing considerable advantages. In the presence of move actions, few-shot learning improves DA Accuracy from 29.04\% to 41.68\%, while in their absence, it rises from 40.82\% to an impressive 54.06\%. We also looked at the selective removal of move acts in a heuristic fashion; for more analysis of the impact of move actions and their selective removal \ref{app:ex:speakornot}


\section{Conclusions and Future Work}

We present an LLM-based user simulator for embodied AI research. Our approach leverages zero-shot and few-shot learning techniques, using LLMs to predict user actions during task-oriented dialogues. The experimental results demonstrate that the proposed method effectively simulates user behavior, achieving a Speak-F1 of 43.4\% and a DA-F1 of 51.13\% in the best-case scenario. This is a significant improvement over traditional baselines, where the user simulator only speaks when spoken to or defaults to a majority class for dialogue acts. We consider this paper as the first step towards building a more comprehensive model. For example, an open LLM like Llama-3 can be fine tuned with the TEACh data. The current model does not incorporate visual information, which is a crucial aspect of embodied AI. Future work could explore integrating visual cues into the user simulator, enabling it to initiate dialogue based on observations of the environment and the agent's actions. For instance, a visual LLM such as GPT-4V or LLaVa could enable the model to determine when to initiate dialogue based on visual cues, like observing the robot wandering around. Furthermore, we plan to plug in this simulator to a state of art embodied agent, such as the HELPER system~\cite{sarch2024helper}, to enable researchers to perform many potential experiments for robot self-play.

\section{Impact Statement}
This work advances the field of embodied AI by demonstrating a scalable and efficient approach to simulate user interactions with embodied agents. The potential applications of this research are wide-ranging, from enhancing task efficiency in human-robot collaboration to providing robust test environments for developing embodied agents that can learn from AI-generated user feedback. Enhanced capabilities in LLM-based agents may inadvertently facilitate misuse, particularly in sensitive domains where autonomous actions could be manipulated for harmful purposes. We urge further exploration into the ethical and security implications of such advanced AI systems to ensure responsible deployment and robust mitigation of associated risks.

\section{Acknowledgments}

This work was supported in part by Other Transaction award HR0011249XXX from the
U.S. Defense Advanced Research Projects Agency (DARPA) Friction for Accountability in Conversational Transactions (FACT) program and has benefited from the Microsoft Accelerate Foundation Models Research (AFMR) grant program, through which leading foundation models hosted by Microsoft Azure and access to Azure credits were provided to conduct the research. This research used the Delta advanced computing and data resource which is supported by the National Science Foundation (award OAC 2005572) and the State of Illinois. Delta is a joint effort of the University of Illinois Urbana-Champaign and its National Center for Supercomputing Applications.

\bibliographystyle{plainnat}
\bibliography{refer}


\newpage
\appendix
\section{Appendix}

\subsubsection{Dialogue act explanation}
\label{app:ex:da-exp}

\begin{table}[ht]
\centering
\begin{tabular}{lll}
\toprule
Dialog Act & Category & Example \\
\midrule
Instruction & Instruction & fill the mug with coffee \\
ReqForInstruction & Instruction & what should I do today? \\
RequestOtherInfo & Instruction & How many slices of tomato? \\
RequestMore & Instruction & Is there anything else to do \\
InfoObjectLocAndOD & Object/Location & knife is behind the sink \\
ReqForObjLocAndOD & Object/Location & where is the mug? \\
InformationOther & Object/Location & Mug is already clean \\
AlternateQuestions & Object/Location & yellow or blue mug? \\
Acknowledge & Generic & perfect \\
Greetings & Generic & hello \\
Confirm & Generic & Should I clean the cup? \\
MiscOther & Generic & ta-da \\
Affirm & Generic & Yes \\
Deny & Generic & No \\
FeedbackPositive & Feedback & great job \\
FeedbackNegative & Feedback & that is not correct \\
OtherInterfaceComment & Interface & Which button opens drawer \\
NotifyFailure & Interface & not able to do it \\
\bottomrule\\
\end{tabular}
\caption{List of dialogue acts from the Dialogue Act Annotation done on TEACh by \cite{Gella2022}.}
\label{tab:dialog-acts}
\end{table}

\subsection{Example Prompt}
\label{app:ex}
Prompts used were made of a few manually written components, and a few randomly generated components.

\subsubsection{Initial Instructions}
\label{app:ex:initial}

{\ttfamily \footnotesize Imagine you, the COMMANDER, are an embodied agent in a simulated world. Your purpose is to instruct a robot, named DRIVER, to do tasks for you by telling it what to do and interrupting it to give further instruction when necessary. Your job here is to predict when you should be giving instructions to the DRIVER based on turn history with the DRIVER. If there is nothing to do or say, you should just observe.}

\subsubsection{Task to the user agent}

{\ttfamily \footnotesize \raggedright \onehalfspacing
Your job is to respond to a given dialogue/action history with only one Dialogue act or OBSERVE. \\
Either return the dialogue act, or return the OBSERVE action. Return only one word/phrase. \\

Goal: Prepare coffee in a clean mug. \\
COMMANDER: <observe> \\
DRIVER: hi,what should i do today?<<Greetings/Salutations,RequestForInstruction>> \\
COMMANDER: Add coffee to a mug <<Instruction>> \\
DRIVER: <observe> \\
COMMANDER: Mug is in the coffee maker already <<InformationOnObjectDetails>> \\
DRIVER: <observe> \\
COMMANDER: <observe> \\
DRIVER: should i rinse the mug or not? <<AlternateQuestions>> \\
COMMANDER: <observe> \\
DRIVER: <toggle on CoffeeMachine> \\
COMMANDER: dont <<Deny>> \\
DRIVER: <observe> \\
COMMANDER: its clean <<InformationOther>> \\
DRIVER: <observe> \\
COMMANDER: <observe> \\
DRIVER: <toggle off CoffeeMachine> \\
COMMANDER: <observe> \\
DRIVER: done <<Acknowledge>> \\
COMMANDER: <observe> \\
DRIVER: what should i do next? <<RequestForInstruction>> \\ 
COMMANDER: the mug doesnt have coffee yet <<InformationOther>> \\
DRIVER: <observe> \\
COMMANDER response:}

\subsubsection{Example answers}
Examples are randomly selected from the data. 

{\ttfamily \footnotesize \raggedright \onehalfspacing
Example : \\
Goal: Prepare breakfast. \\
COMMANDER: <observe> \\
DRIVER: hello what are my tasks <<Greetings/Salutations,RequestForInstruction>> \\
COMMANDER: hii <<Greetings/Salutations>> \\
DRIVER: <observe> \\
COMMANDER: prepare coffe in clean mug <<Instruction>> \\
DRIVER: <observe> \\
COMMANDER: <observe> \\
DRIVER: <pickup Mug> \\
COMMANDER: <observe> \\
DRIVER: <putdown CounterTop> \\
COMMANDER response: \\
OBSERVE}

\subsubsection{Variations}
Variation of the dialogue history without Driver <move> acts:

{\ttfamily \footnotesize \raggedright \onehalfspacing
Goal: Put all Book on any Furniture.\\
COMMANDER: <observe>\\
DRIVER: hello how may i help \\<<Greetings/Salutations,RequestForInstruction>>\\
COMMANDER: hi <<Greetings/Salutations>>\\
DRIVER: <observe>\\
COMMANDER: put the cook on furniture <<Instruction>>\\
DRIVER: <observe>\\
COMMANDER: book <<Instruction>>\\
DRIVER: <observe>\\
COMMANDER: <observe>\\
DRIVER: <pickup Book>\\
COMMANDER: <observe>\\
DRIVER: <putdown Desk>\\
COMMANDER: the book is in the small room <<InformationOnObjectDetails>>\\
DRIVER: <observe>\\
COMMANDER: <observe>\\
DRIVER: <pickup Book>\\
COMMANDER: <observe>\\
DRIVER: <putdown Desk>\\
COMMANDER: <observe>\\
DRIVER: done <<Acknowledge>>\\
COMMANDER: <observe>\\
DRIVER: how else can i help you <<RequestMore>>\\
COMMANDER response:}

\subsection{To Speak or Observe? Impact of Action Dynamics on GPT-4’s User Behavior Simulation}
\label{app:ex:speakornot}
The confusion matrices illustrate the performance of GPT-4 in predicting when to speak and when to observe across four different experimental conditions, simulating user behavior in a human-robot interaction scenario. In the zero-shot setting with move actions, the model demonstrates a high rate of false positives (2107 instances), indicating a tendency to speak excessively even when it should be observing. This suggests that the model, without prior training examples, struggles to balance its actions, leading to inappropriate conversational behavior. Comparatively, in the zero-shot without move actions condition, there is a noticeable improvement in model performance, with a reduction in false positives (1408 instances) and an increase in true positives (752 instances). This indicates that the absence of move actions reduces the complexity of the decision-making process, allowing the model to identify appropriate moments to speak more accurately.
\begin{figure*}[ht]
\centering
\includegraphics[width=0.95\textwidth]{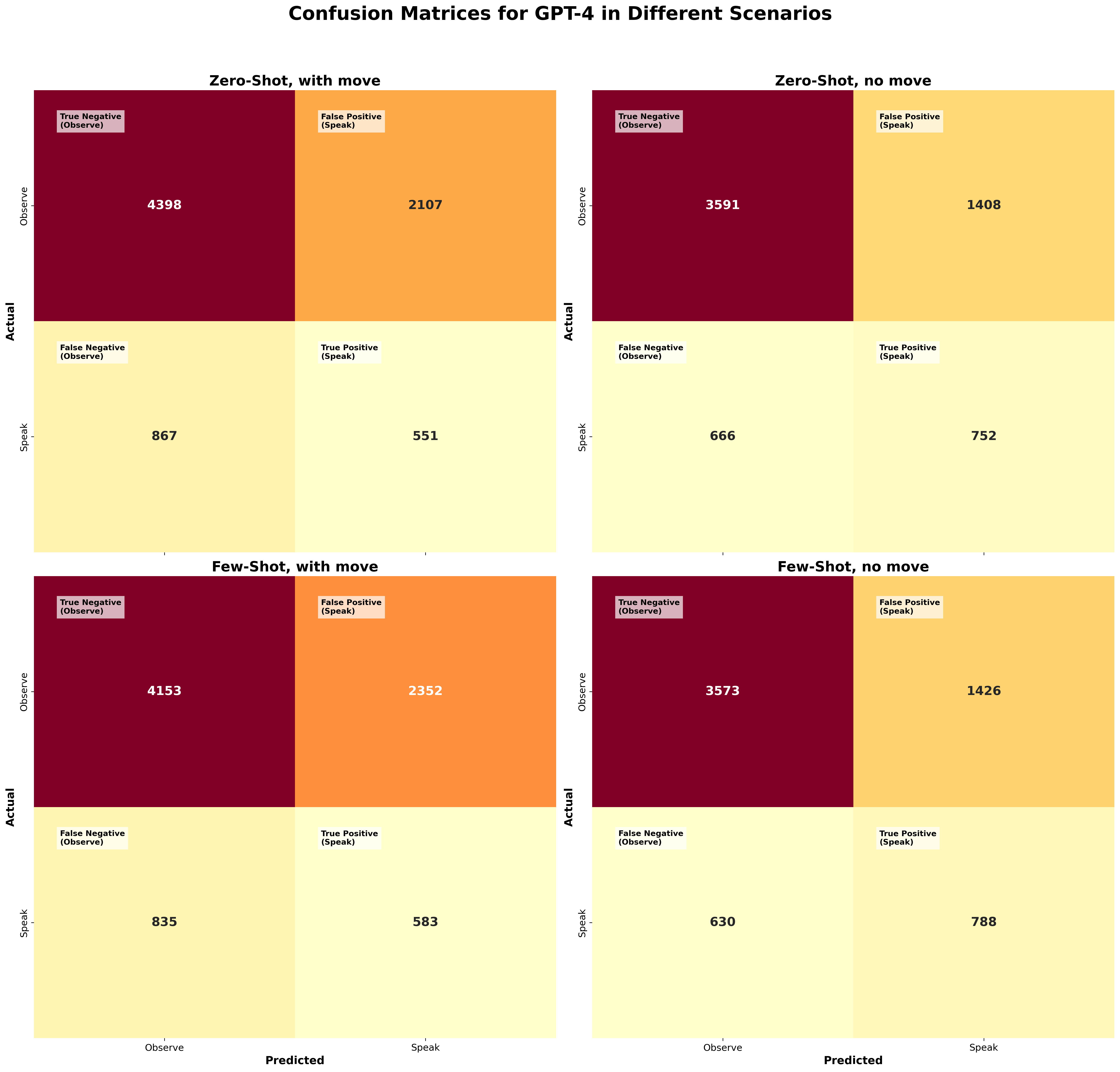}
\vspace*{-2ex}
\caption{Confusion Matrices illustrating GPT-4’s performance in simulating user behavior across four conditions: zero-shot and few-shot learning with and without move actions.}
\label{fig:conf-mat-da}
\vspace*{-2ex}
\end{figure*}
When transitioning to the few-shot learning scenarios, the model shows mixed improvements. In the few-shot with move actions condition, while there is a slight increase in true positives (583 instances), the number of false positives (2352 instances) remains high, reflecting continued over-talkativeness. This suggests that few-shot learning alone may not be sufficient to mitigate the confusion introduced by move actions. On the other hand, in the few-shot without move actions condition, the model achieves its best performance, with the highest true positive rate (788 instances) and the lowest false negative rate (630 instances). This demonstrates that the combination of few-shot learning and the removal of move actions enables the model to better emulate user behavior, accurately deciding when to speak and when to remain silent.
\subsubsection{Selective Removal of Move Actions}
To see if move actions provided meaningful context to the LLM despite often causing noise, we added a method of selective removal of move actions. The criteria for removing a move turn was whether it directly followed a question asked by the robot. In this case, the robot was expecting a response from the user, and in most cases the user responded. Often, however, the TEACh dataset contained a move turn directly following these questions, most likely caused by simple human error, but when the LLM sees this move, it decides not to answer after that, causing an error in evaluation for two moves.
\begin{table*}[ht]
    \centering
    \begin{tabular}{l|c|c|c|c}
    \toprule
    & \multicolumn{2}{c|}{\textbf{Speak-F1}} & \multicolumn{2}{c}{\textbf{DA-F1}} \\
    \midrule
    \textbf{Approach} & Moves Excluded & Selective Removal & Moves Excluded & Selective Removal \\
    \midrule
    Zero-shot & 0.344 & 0.435 & 0.292 & 0.433 \\
    Few-shot  & 0.399 & 0.537 & 0.361 & 0.528 \\
    \bottomrule
    \end{tabular}
    \caption{F1-measures for Speak and Dialogue Act metrics for GPT-4, run on the dataset without moves and the dataset with moves selectively removed.}
    \label{tab:selective-removal}
\end{table*}

Results from running on this modified dataset were mostly similar to excluding the move turns entirely, but in some cases the selective removal performed noticeably worse than before. Thus, the move turns do not provide meaningful information to the user, most likely due to a lack of environmental context.

\end{document}